\documentclass[]{interact}

\usepackage{epstopdf}

\usepackage[numbers,sort&compress]{natbib}
\bibpunct[, ]{[}{]}{,}{n}{,}{,}
\renewcommand\bibfont{\fontsize{10}{12}\selectfont}
\makeatletter
\def\NAT@def@citea{\def\@citea{\NAT@separator}}
\makeatother

\theoremstyle{plain}

\theoremstyle{definition}

\theoremstyle{remark}

\usepackage{url}
\usepackage{array,tabularx}
\usepackage{multicol} 
\usepackage{algorithm}
\usepackage{algorithmic}
\usepackage{caption}
\usepackage{subcaption}
\usepackage[dvipsnames]{xcolor}
\usepackage{comment}
\usepackage{hyperref}
\usepackage{authblk}

\begin{document}

\articletype{Research Article}

\title{Vision-based indoor localization of nano drones in controlled environment with its applications}
\author[1]{Simranjeet Singh}
\author[2]{Amit Kumar}
\author[2]{Fayyaz Pocker Chemban}
\author[2]{Vikrant Fernandes}
\author[1]{Lohit Penubaku}
\author[3]{Kavi Arya}
\affil[1]{Electrical Engineering Department, Indian Institute of Technology Bombay, Mumbai, India}
\affil[2]{Embedded Real-Time Systems {/} e-Yantra Lab, Indian Institute of Technology Bombay, Mumbai, India}
\affil[3]{Computer Science and Engineering Department, Indian Institute of Technology Bombay, Mumbai, India}


\maketitle

\begin{abstract}
Navigating unmanned aerial vehicles in environments where GPS signals are unavailable poses a compelling and intricate challenge. This challenge is further heightened when dealing with Nano Aerial Vehicles (NAVs) due to their compact size, payload restrictions, and computational capabilities. This paper proposes an approach for localization using off-board computing, an off-board monocular camera, and modified open-source algorithms. The proposed method uses three parallel proportional-integral-derivative controllers on the off-board computer to provide velocity corrections via wireless communication, stabilizing the NAV in a custom-controlled environment. Featuring a 3.1cm localization error and a modest setup cost of 50 USD, this approach proves optimal for environments where cost considerations are paramount. It is especially well-suited for applications like teaching drone control in academic institutions, where the specified error margin is deemed acceptable. Various applications are designed to validate the proposed technique, such as landing the NAV on a moving ground vehicle, path planning in a 3D space, and localizing multi-NAVs. The created package is openly available at \href{https://github.com/simmubhangu/eyantra_drone}{https://github.com/simmubhangu/eyantra\_drone} to foster research in this field.
\end{abstract}


\section{Introduction}\label{sec1}
For thousands of years, humans have used land and water to navigate terrain, but air travel has revolutionized many aspects of our lives. Recent advances in materials and electronics have led to the development of portable aerial robots, which can be either user-operated or autonomous. Research is underway to enhance the autonomous capabilities of these unmanned aerial vehicles (UAVs), classified as fixed-wing, rotating \& flapping wing, hybrid wing, and gas envelope~\cite{Lee2021}. Among these, quadcopters, a type of rotating \& flapping wing aircraft, are widely used and have numerous applications in disaster management, defense, cinematography, agriculture, and beyond. UAVs can also be categorized by size, ranging from nano to large, depending on their dimensions~\cite{classProgInAero}. As we transition from the nano to larger drones, both computing power and costs begin to rise.

However, nano aerial vehicles (NAVs) have garnered significant interest in educational research and initial testing of aerial algorithms due to their potential for autonomous maneuvering in indoor and outdoor environments. Since global positioning systems (GPS) are not available indoors, localization of NAV in the environment becomes a crucial challenge. To navigate the surroundings, NAV must localize themselves in the environment by obtaining their $3D$ pose with respect to the environment and heading angle. Localization using internal sensors, such as an inertial measurement unit, can be robust for a short duration but is prone to drifting in $3D$ space when used for longer periods~\cite{imu_paper}. Additionally, NAVs are small and have limited payload capacity, making it challenging to accommodate various sensors. As a result, researchers have explored alternative localization methods for indoor drone applications.

 \begin{figure}[!t]
    \centering
    \includegraphics[width=0.8\linewidth]{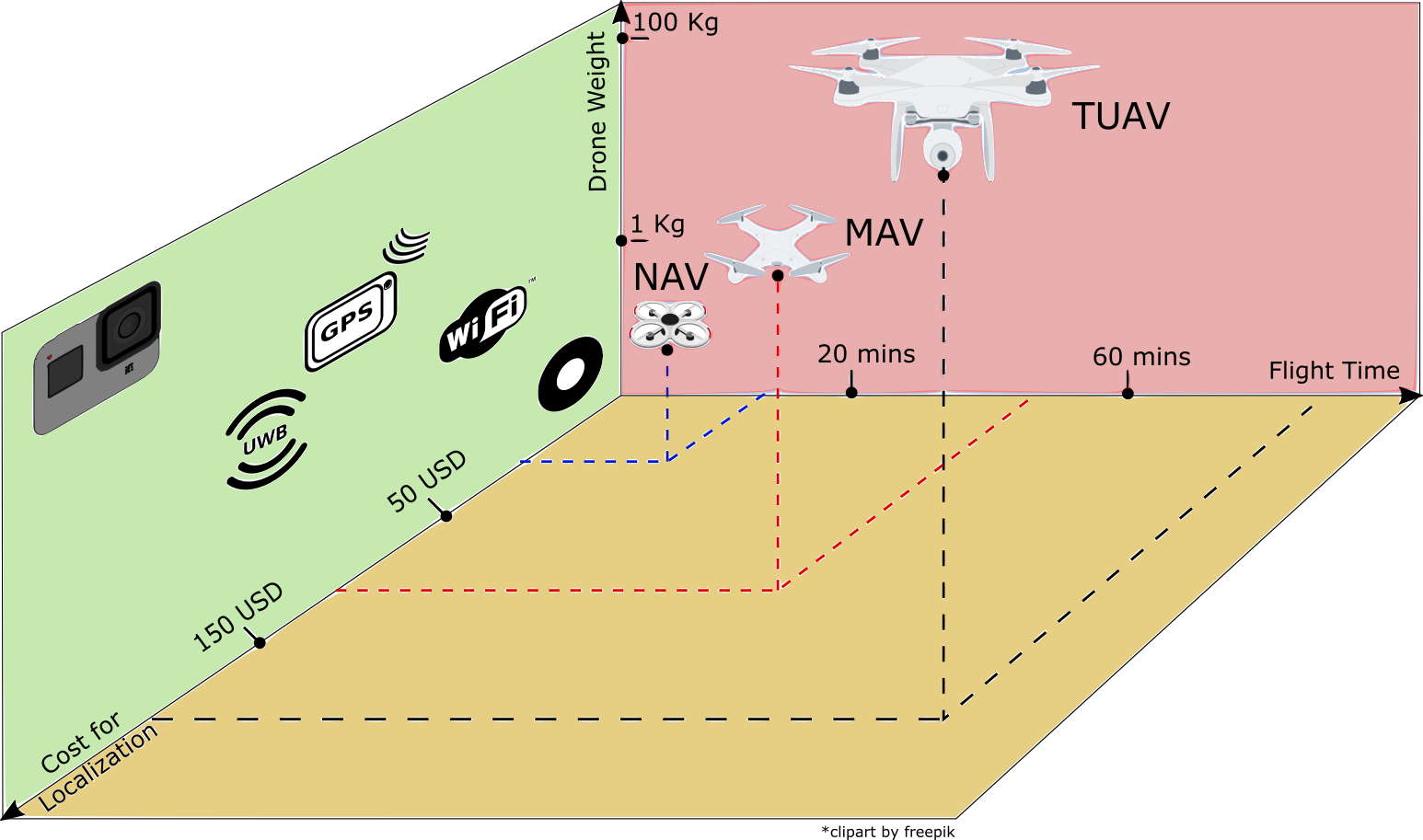}
    \caption{Comparison between flight time, drone weight, and cost for localization of nano aerial vehicle (NAV), miniature aerial vehicle (MAV) and tactical unmanned aerial vehicle (TUAV). Using NAV along with WhyCon marker for localization, substantially reduces the setup cost.}
\label{fig:why_our_paper}
\end{figure}

To tackle the localization problem in an indoor environment, various techniques have been proposed to develop indoor positioning systems by adding artificial landmarks or beacons at known locations such as ViCon~\cite{vicon}, radio frequency-based (RF) localization system~\cite{kempke2015polypoint}, and distributed sensor systems~\cite{nam2017unmanned}. Even though these localization techniques achieve excellent accuracy and performance, they are costly and only affordable for some research fields. Figure~\ref{fig:why_our_paper} compares the cost of different localization techniques with different types of drones. Researchers have also developed techniques for indoor localization using vision-based approaches. These approaches employ fiducial markers such as AprilTags~\cite{wang2016apriltag}, ArUco markers~\cite{arucoopencv} or circular markers~\cite{nitsche2015whycon} and an onboard camera. They are capable of position estimation of the marker in the camera video stream. However, these approaches demand an onboard computational sensor processing unit for such localization algorithms. This is elevated by sending the sensor data to the ground station to run a computation-heavy localization algorithm~\cite{Onboard_IMU_monocular_vision}. However, the ground station approach suffers from the latency and communication gap between NAV and the ground station. 


A viable solution is to localize the NAV using a vision-based system. Recently, WhyCon, an efficient, fast, and low-cost open-source localization system, has been presented~\cite{nitsche2015whycon}. It uses off-the-shelf components and circular markers for localization. This system requires no special equipment to set up the controlled environment. A lightweight and low-cost robotics swarm implementation using the WhyCon system is presented in~\cite{lowcost}. It has been shown that the WhyCon localization system outperforms ArUco and OpenCV by a hundred times in terms of processing time~\cite{whyconjint}. However, the foundational WhyCon system does not provide a unique ID of the target pose, making it difficult for fast control loop applications such as swarm NAVs~\cite{Fast_Fiducial_Marker}.

The major contributions of this paper are as follows:
\begin{itemize}
    \item Design of a localization-controlled environment for NAV using a WhyCon system (open-source) that is low-cost, easy to set up, and accurate, as illustrated in Figure~\ref{fig:system_intro}.
    \item Design of complete architecture of localization system, control commands, and communication protocol around~NAVs. 
    \item Validation of the proposed localization system by implementing applications such as landing on the objects in motion, autonomous path-planning, and multi-NAV control.
    \item Public release of the package developed to maneuver the NAV in the controlled environment at \\ 
    \url{https://github.com/simmubhangu/eyantra_drone}
\end{itemize}

While all the individual components of our system may not be novel, to our knowledge, our novelty lies in the integration of a complete system that has been tested on various applications suitable for a low-cost academic environment. The rest of the paper is organized as follows: Section~\ref{related_work} reviews the existing and related work on vision-based localization systems for NAVs, Section~\ref{enviornment_setup} provides the details of the localization setup and experiments, Section~\ref{control_system} presents the control architecture for NAV, and Section~\ref{application} lists the various applications designed on the proposed technique.

 \begin{figure}[!t]
    \centering
    \includegraphics[width=0.4\textwidth]{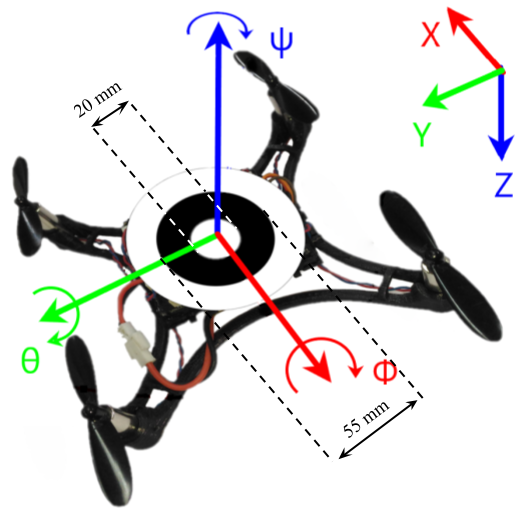}
    \caption{Coordinate frames of the NAV and overhead camera. WhyCon marker, shown with an inner white circle and black boundary, is mounted on top of the NAV. Changes in roll ($\phi$), pitch ($\theta$), and throttle result in movement along the y, x, and z axes respectively.} 
\label{fig:drone_cord}
\end{figure}

\section{Background and Related Work}\label{related_work}

In this section, we delve into the related work and background of the components employed in the proposed architecture, including the NAV, state-of-the-art vision-based techniques, and the control algorithm designed for NAV. 

\subsection{Related Work}
There have been several implementations of indoor localization systems in the literature. Some of the early work focused on placing artificial landmarks~\cite{Vicon-460,beacon} in an unknown environment for object tracking. Even though Vicon-460~\cite{Vicon-460} system provides an overall accuracy of $65\pm5 \mu m $, these approaches remain costly and are not viable for every research field. This issue has motivated researchers to look into low-cost localization systems. As an alternative, various marker-based localization techniques have been developed~\cite{wang2016apriltag,arucoopencv, nitsche2015whycon, Aruco}. These markers consist of different patterns; some encode data into them. These patterns are detected using vision algorithms, which provide the pose, angle, and information encoded in it if it exists. A comparative analysis of marker-based localizations has been outlined in~\cite{whyconjint}, revealing that the WhyCon marker stands out for its speed and reduced computational demands~\cite{nitsche2015whycon,whycon_icar}. This advantage is attributed to its straightforward encoding and tracking technique.

All the localization techniques mentioned above can localize a UAV in the environment. Different variations of these techniques have been proposed in the literature. For example, researchers were pasting the fiducial markers on the wall at a known location and localizing the UAV using onboard vision processing or relative localization~\cite{Low_cost_embedded,MAV_top}. The implementation of indoor localization using an onboard camera and computing is presented in~\cite{rgbd}. The implementation in~\cite{whycon_icar} requires an onboard camera, onboard computing, and a fixed fiducial marker on the wall. However, these techniques require a high payload-carrying capacity for sensory elements, which is difficult for NAVs.


Many low-cost techniques for vision-based localization have been presented in the literature, such as maintaining a geometric pattern visible to another aerial vehicle~\cite{Low_cost_embedded}. This approach is straightforward and relies solely on off-the-shelf components. However, it necessitates on-board computing, restricting processing capabilities. Alternatively, other vision-based techniques utilizing stereo vision have been proposed to alleviate on-board computation~\cite{Stereo_vision}. The architecture suggested in~\cite{Stereo_vision} computes real-time 3D coordinates, providing this feedback to the UAV. Yet, this method involves a comprehensive stereo-vision preprocessing pipeline, limiting the update rate to 10Hz. Following this, a detailed breakdown of the key components is presented individually.

\begin{figure}[!t]
  \centering
  \includegraphics[width=0.8\linewidth]{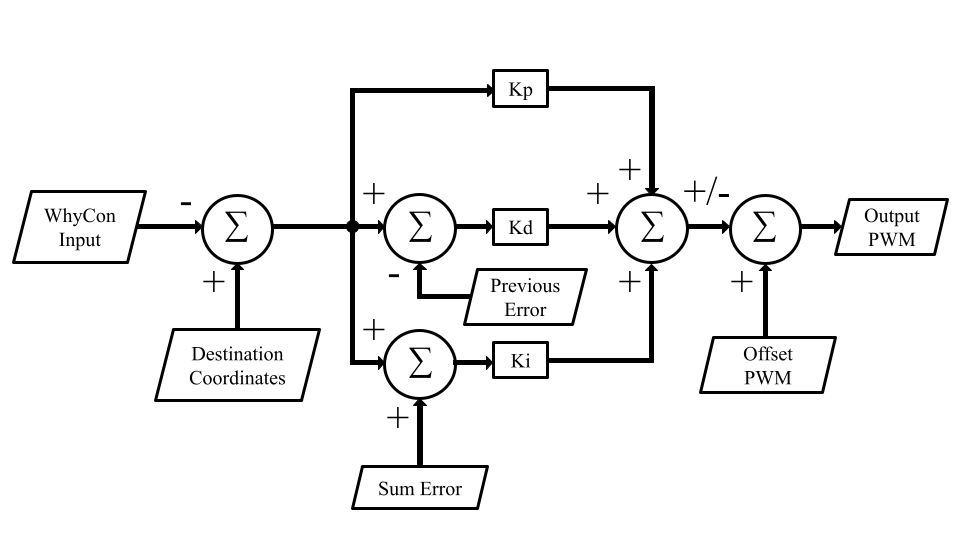}
  \caption{Block diagram of external PID-Controller used for roll ($\phi$), pitch ($\theta$), and throttle. These controllers work in parallel to stabilize the drone.}
\label{fig:pid_block_dia}
\end{figure}

\subsection{Nano Aerial Vehicle (NAV)}
Unmanned aerial vehicles are categorized based on payload capacity and flight time. As illustrated in Figure~\ref{fig:why_our_paper}, various aerial vehicles are compared in terms of flight time and localization cost. Given the focus of this study on designing an affordable localization system for low-cost drones, the investigation is confined to NAV.

The NAV utilized in this experiment is ``Pluto X," an open-source drone created by Drona Aviation~\cite{Drona}, a startup affiliated within the institution. The Pluto X boasts compact dimensions of 9x9 cm, a lightweight profile at 48 grams, and a flight endurance of 7 minutes. It serves as an exceptionally versatile platform for developers. Powered by a 3.7V 600mAh LiPo battery, it can achieve a maximum speed of 3 m/s. Its motor controls facilitate dynamic translation along the $x$, $y$, and $z$ axes, as well as rotation about the roll ($\phi$), pitch ($\theta$), and yaw ($\psi$) axes in three-dimensional space, as illustrated in Figure~\ref{fig:drone_cord}.


The drone has an onboard set of sensors, including a three-axis accelerometer, gyroscope, magnetometer, barometer, and a bottom-mounted laser sensor for height determination. These sensors are internally filtered to provide essential data on roll, pitch, yaw, and altitude. Additionally, the drone features onboard Wi-Fi for bidirectional data exchange and firmware programming directly on the device.

  


 
\subsection{Indoor Localization}
The pose data from WhyCon can be used as feedback for external control algorithms. The authors in ~\cite{nitsche2015whycon} show that the proposed algorithm in WhyCon marker detection can find patterns approximately one thousand times faster than traditional methods. The fast update rate of the WhyCon algorithm, 80 to 100 Hz, made it suitable for obtaining precise and high-frequency localization information for our control loop to maintain the reference drone pose. Furthermore, the high update rate allowed us to track multiple markers without delaying the control loop of the drone. The WhyCon marker is made of an inner white patch surrounded by a black portion, as shown in Figure~\ref{fig:drone_cord}. The inner diameter can be reduced or increased. However, changing it affects the $z$ coordinate.

However, the original design of the WhyCon localization algorithm assumes a fixed number of targets within the frame, posing a challenge when endeavoring to change these targets dynamically. This calls for an adaptation of the current algorithm to adeptly handle dynamic targets within the frame.


\subsection{Control Algorithm}
Controllers are designed to produce commands that direct a plant, in this instance, NAV, to reach a specified setpoint within a finite time frame. One such controller, Proportional Integral Derivative (PID) controller is widely adopted in drones~\cite{why_pid}. It is a negative feedback control mechanism used to design a system that automatically applies an accurate and responsive correction to a control function and stabilizes at the desired output or reference. As the NAV has four controllable degrees of freedom, we must apply the PID controller to all of them, i.e., $\phi$, $\theta$, $\psi$, and throttle as shown in Figure~\ref{fig:pid_block_dia}. However, the NAV is placed in its starting position so that its yaw is at the desired setpoint with respect to the overhead camera. The internal PID of the NAV makes sure that the yaw remains the same throughout the flight. To this end, we must apply a parallel PID control system to control $\phi$, $\theta$, and throttle. 

\section{Proposed Localization Setup and Experiments}\label{enviornment_setup}

This section details all the crucial aspects of the proposed localization setup and experiment performed. In subsection \ref{cus_con_envir}, details about the environment such as the camera, platform, and drone used are discussed, in subsection \ref{Internal and External Control}, internal controller of the NAV is discussed, in subsection \ref{local_whycon_marker} localization method using a WhyCon marker is explained, and in subsection \ref{camera_callibration}, the process used for calibrating the camera is mentioned.

 \begin{figure}[!t]
    \centering
    \includegraphics[width=0.5\linewidth]{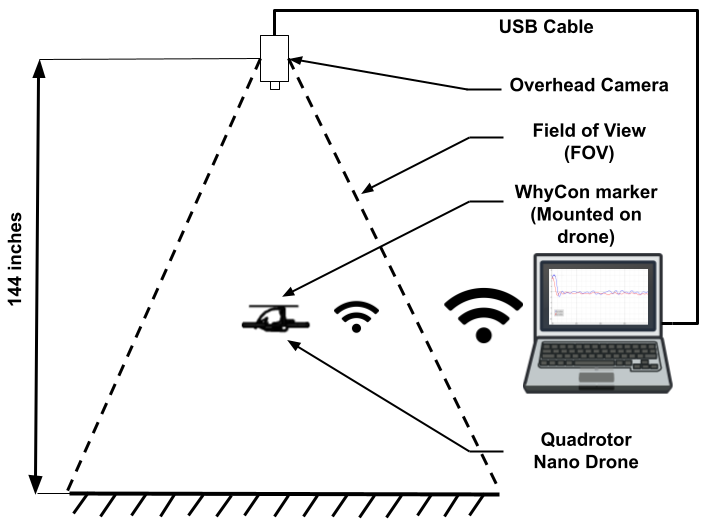}
    \caption{Arrangement of a custom-controlled localization environment for NAV utilizing an adapted WhyCon system.}
\label{fig:system_intro}
\end{figure}

\subsection{Custom-Controlled Environment}
\label{cus_con_envir}




The setup of indoor controlled environment is shown in Figure~\ref{fig:system_intro}. Overhead camera, ELP 2 Mega Pixel Ov2710, is mounted at 3-5 meters. It offers a wide field of view adequate for the NAV to maneuver. The camera has a maximum resolution of 1920x1080 and a  maximum FPS of 120. Additionally, the camera provides a large field of view (FOV) of 170 degrees to view most of our controllable environment. As the region enclosed in the controlled environment depends on the height of the camera, the resolution of the frame, and the view angle of the camera, the resolution is set to 640x480 at 120 FPS. A high FPS ensures faster localization information for the control algorithm.


Robot Operating System (ROS), running on the laptop shown in Figure~\ref{fig:system_intro}, is a framework consisting of various libraries and packages that can reduce the complexity of tasks. ROS provides high flexibility allowing developers to test their work in simulators before working on robot hardware. It provides topics and services to share the data amongst different live nodes.

The drone's firmware is built upon the open-source firmware Cleanflight~\cite{cleanflight}, which provides the support of Multiwii Serial Protocol (MSP)~\cite{MSP}. Onboard firmware is modified so that it is capable of communicating through MSP via  Wi-Fi. Drone-ROS Driver (DRD) is developed for handling the communication of drones with ROS~\cite{rosdriver}. MSP is a simple and light protocol that communicates coded messages. The messages' format, direction, and ID are unique and predefined.


\begin{figure}[!t] 
\centering
\includegraphics[width=0.8\linewidth]{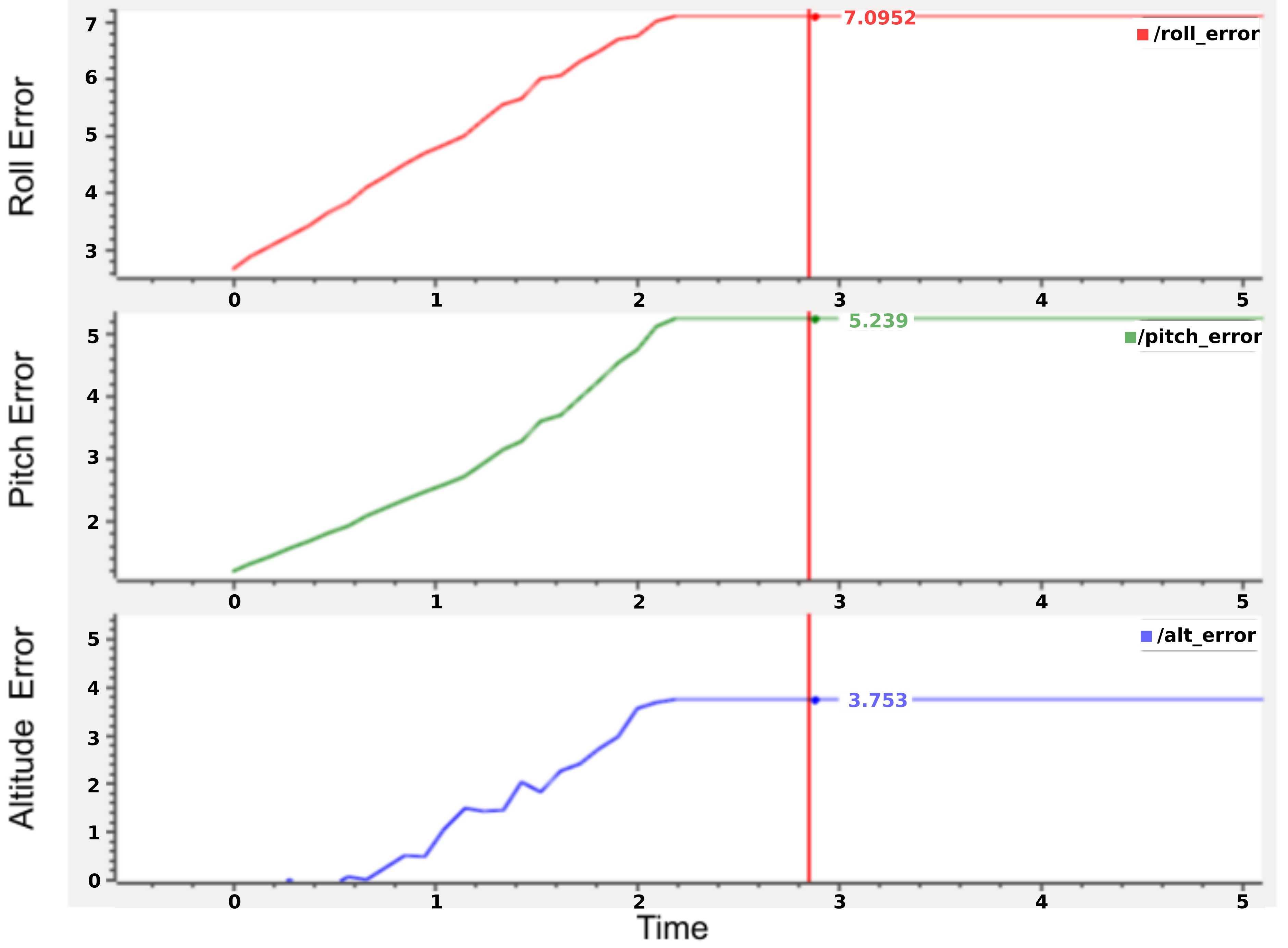}
\caption{$\phi$, $\theta$ and Altitude error with internal control. As depicted in the graph, there is an evident increase in error of roll, pitch, and altitude with time. This indicates the necessity for an external controller.}
\label{manual_error}
\end{figure} 


Drone data is shared through ROS service with a dedicated message type. ROS topic- `drone\_command,' has the Pulse Width Modulation (PWM) values for maneuvering the drone. Nine messages can be passed to the drone for control - four AUX (Auxiliary) channels, i.e., $rcAUX1, rcAUX2$, $rcAUX3$ \& $rcAUX4$, four control arguments, i.e., $rcRoll$, $rcPitch$, $rcYaw$ \& $rcThrottle$, and $droneIndex$ to select the drone index for multi drone control. AUX channels are used to arm \& disarm the drone and control other modes, such as ``THROTTLE MODE" or ``ALT HOLD MODE." The default value of these parameters is 1500 and can be varied from 1000 to 2000. Onboard sensors provide data about $\phi$, $\theta$, $\psi$, acceleration, altitude, battery, and RSSI (Received Signal Strength Indication) via ROSService. DRD can control multiple drones at a time using the same message. $droneIndex$ is used to identify the ID of the drone on which the message is to be delivered.



\subsection{Internal and External Control}
\label{Internal and External Control}
The NAV has an internally filtered IMU sensor which is used to regulate the $\phi$, $\theta$, and $\psi$ of the drone. It is used to hold the NAV steady at its own axis. We measured its ability to hold the position at a fixed setpoint in terms of WhyCon coordinates with and without our external control algorithm. The error measured is the difference between the setpoint and the drone's current position in WhyCon coordinates. 

\begin{figure}[!t] 
\centering
\includegraphics[width=0.7\linewidth]{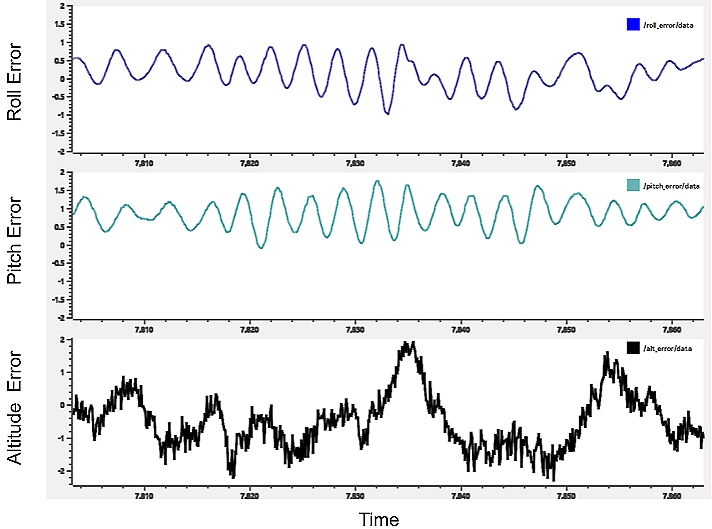}
\caption{$\phi$, $\theta$ and Altitude error with both internal and external control. It is evident from the above graph that the addition of an external controller is required along with the internal sensors of the NAV.}
\label{pid_error}
\end{figure}

As shown in Figure \ref{manual_error}, the NAV starts at the setpoint but drifts off in all three axes until it is completely out of the frame. The graph flat-lines once it is out of the frame. Maximum error recorded in $\phi$ is 7.095, $\theta$ is 5.23, and altitude is 3.75 WhyCon coordinates. An external control prevents the drone from drifting out of the frame. From Figure \ref{pid_error}, we found that the error in position hold is quite small. Maximum recorded error in $\phi$ is between -0.9 and 0.8, in $\theta$ is between -0.1 and 1.8, and in altitude is between -2.2 and 2. The comparison results show that the NAV itself cannot hold the position without external control. Its onboard sensors are inaccurate and hence fail to localize itself.

\subsection{Localization using WhyCon Marker}
\label{local_whycon_marker}
To obtain the 6 degrees of freedom (DOF) pose of the drone in the real world, we used an external camera to track the drone. The WhyCon visual marker was mounted on the drone to provide this. Although the orientation data is unreliable, the position data from WhyCon is extremely accurate. As shown in Figure \ref{fig:drone_cord}, to express the drone coordinate frame (D) in the camera coordinate frame (C), a 180$^\circ$ 3D rotation transform was applied across the $y-$axis. 

In general, let $^\textup{C}T_D$ be the homogeneous transform from the drone coordinate frame (D) to the camera coordinate frame (C). It can be expressed as:

\begin{center}
$
\begin{vmatrix}
^\textup{C}R_D&^\textup{C}t_D\\
0&1\\
\end{vmatrix}
$
\end{center}
where, $^\textup{C}R_D$ is the rotation matrix and $^\textup{C}t_D$ is the translation matrix. Applying a $180^\circ$($\theta$) 3D rotation across the $y-$axis, we represent our $^\textup{C}T_D$ matrix as:

\begin{center}
$
\begin{vmatrix}
cos\theta&0&sin\theta&t_x\\
0&1&0&t_y\\
-sin\theta&0&cos\theta&t_z\\
0&0&0&1
\end{vmatrix}
$
\end{center}


\subsection{Camera Calibration}
\label{camera_callibration}
The general image view of USB cameras has a fish-eye view output\cite{Cattaneo_2015}. However, this is undesirable as it affects the WhyCon coordinates. We need a flat image output from our camera, and hence it is necessary to calculate its parameters. Fish-eye radial distortion has two types - Negative radial distortion or Pincushion distortion and Positive radial distortion or Barrel distortion. In either case, the camera's intrinsic and extrinsic parameters are used to map the 3D coordinates of an object in the world frame to the 2D coordinates of the image plane generated. The extrinsic parameters of the camera represent the camera's location in 3D space. It consists of rotation and translation components. These transform an object's 3D world coordinates to the camera's 3D coordinates. The intrinsic parameters of the camera consist of focal length and optical center. Via a projective transform, the 3D camera coordinates are transformed into 2D image coordinates. We can calibrate the camera by determining the extrinsic and intrinsic parameters using printouts of a pattern of defined size, for instance, a checkerboard.
ROS provides support for calibrating a wide variety of cameras, provided that the camera's drivers satisfy the ROS camera interface. Using the camera calibrator package, both monocular and stereo cameras can be calibrated. The package is built over OpenCV camera calibration \cite{camcal} using the transformation as given in Equation~\ref{opencv_package} where $s$ is the scale factor, $(x,y)$ are the image points, $(X,Y,Z)$ are the real world coordinates, $(R,t)$ are the extrinsic parameters of the camera and $K$ is the intrinsic parameter of the camera.

\begin{equation}
\label{opencv_package}
s
\begin{vmatrix}
x&y&1\\
\end{vmatrix}
=
\begin{vmatrix}
X&Y&Z&1\\
\end{vmatrix}
\begin{vmatrix}
R\\
t\\
\end{vmatrix}
K
\end{equation}

\section{Control Architecture}\label{control_system}
\subsection{Position holding of NAV}
The WhyCon marker input consists of $x$, $y$, and $z$ coordinates which gives the current position of the NAV with respect to the world in WhyCon's coordinate frame. These coordinates will be used as feedback to maintain the position of the drone at a particular point. Destination coordinates or the coordinates at which the drone must hold its position are denoted by $\bar{X}$, $\bar{Y}$, and $\bar{Z}$. Error ($x-\bar{X}$, $y-\bar{Y}$, $z-\bar{Z}$) between the current position and the destination is fed to the PID controller as shown in Figure \ref{fig:pid_block_dia}. PID controller takes errors as inputs and generates the expected PWM of the motors so that it can move toward the destination coordinates. These errors are used to control the $\phi$, $\theta$, and throttle of the drone respectively. An internal PID of NAV is running independently to stable it at its own references as mentioned in subsection \ref{Internal and External Control}. A proportional controller ($K_p$) reduces the rise time and gets the drone to oscillate around a point. If that point is not the setpoint, it indicates that the system has a steady-state error. The integral controller ($K_i$) eliminates the steady-state error. A derivative controller ($K_d$) increases the stability of the system by reducing the overshoot.

All three PID controllers take new data after a fixed time interval called sample or loop time. Sample time depends upon the worst-case delay to calculate the output of PID in each iteration as given in Equation \ref{eq:sample_time}. \\
\begin{equation}
    Sample time(ST) = \frac{1}{FPS}+ PID_{time} + buffer
    \label{eq:sample_time}
\end{equation}



\subsection{Waypoint Navigation of NAV}
\begin{figure}[!t] 
\centering
\includegraphics[width=\linewidth]{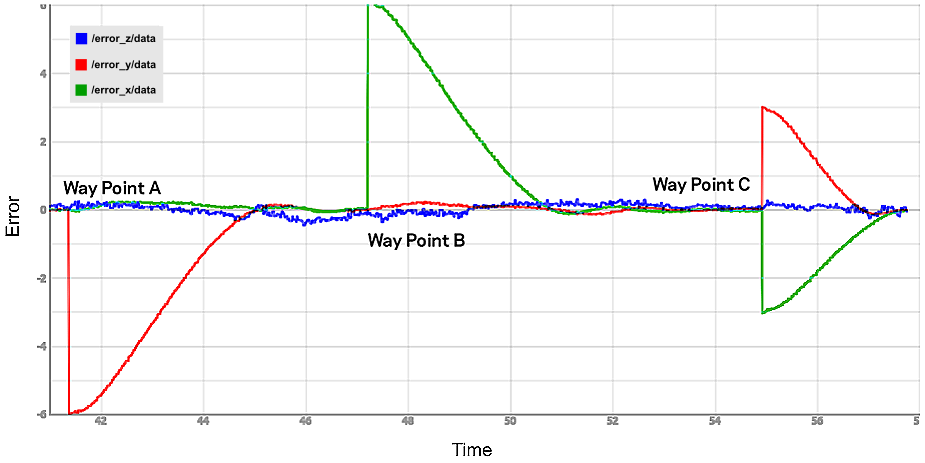}
\caption{Plot of error in y (red), x (green), and z (blue) axes in WhyCon coordinate frame vs time in seconds. The graph indicates the performance of the tuned PID controller. It is evident that there is minimum overshoot and steady-state error for each waypoint.}
\label{Waypoint_navigation}
\end{figure}

Waypoint is a destination coordinate at which the drone must hold its position. Figure \ref{Waypoint_navigation} shows the plot of error in y (error\_y/data), x (error\_x/data), and z (error\_z/data) axes in WhyCon's coordinate frame vs time at different waypoints.

In Figure \ref{Waypoint_navigation}, A denotes the point at which a new waypoint (keeping the $\bar{Z}$ and $\bar{X}$ same and a different $\bar{Y}$) is set. We can see that the rise time is less for the NAV to stabilize at the new setpoint. There is minimum overshoot and steady-state error, indicating optimum tuning of PID parameters in pitch. B denotes the point at which a new waypoint with a different $\bar{X}$ is set, and C denotes the point at which both $\bar{X}$ and $\bar{Y}$ are changed. Factors like rise time, overshoot, settling time, and steady-state error in the plot indicate appropriate tuning of PID parameters in roll and pitch. Similarly, PID parameters for throttle are also tuned.

\subsection{Auto-tuning of PID}
The method of computing the PID parameters using trial and error is both tedious and time-consuming. Since it involves human intervention, the controller may not be optimally tuned. Auto-tuning of controllers can help to solve the above problems. We will discuss Zielger-Nichols Method and Iterative Feedback Auto-tuning in subsections \ref{Zielger-Nichols Method Section} and \ref{Iterative Feedback Auto-tuning Section} respectively.

\subsubsection{Zielger-Nichols Method}
\label{Zielger-Nichols Method Section}
Many techniques have come up over the years to auto-tune the PID controller after the works of Ziegler and Nichols\cite{ziegler1942optimum}. It involves relay feedback\cite{aastrom1984automatic} and pattern recognition\cite{shinskey1994feedback} techniques. We follow a similar approach as that of \cite{luo1998new} in which we induce sustained oscillations and measure PID parameters using Zielger-Nichols reaction curve method. 

\begin{figure}[!t] 
     \centering
     \begin{subfigure}{0.46\textwidth}
         \centering
         \includegraphics[width=\columnwidth]{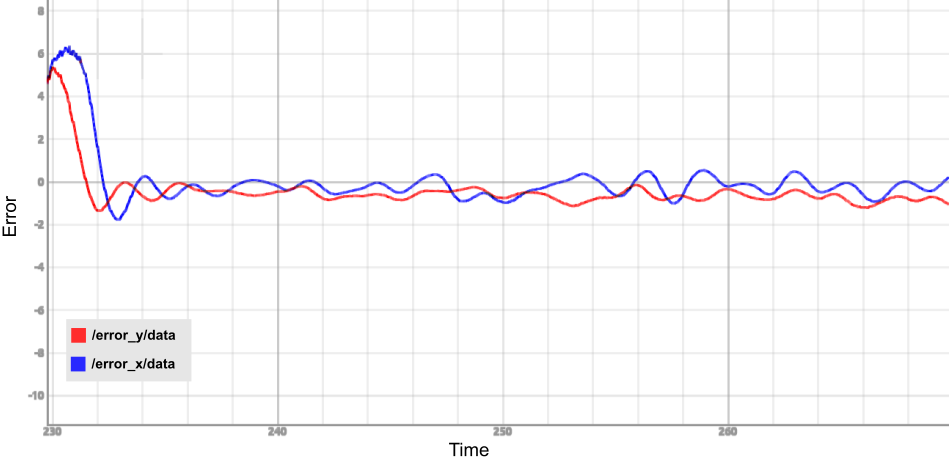}
         \caption{Classic PID controller}
         \label{fig:Classic PID controller}
     \end{subfigure}
     \hfill
     \begin{subfigure}{0.46\textwidth}
         \centering
         \includegraphics[width=\columnwidth]{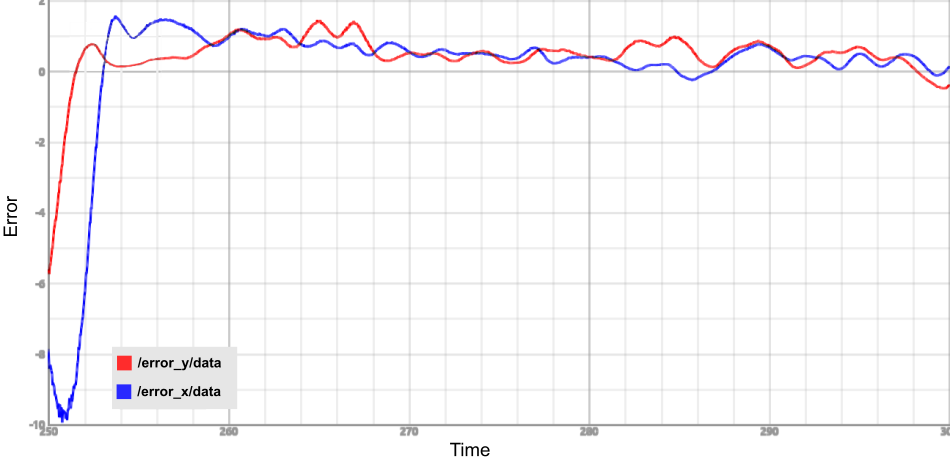}
         \caption{Pessen Integral Rule PID controller}
         \label{fig:Pessen Integral Rule PID controller}
     \end{subfigure}
     \vfill
     \begin{subfigure}{0.46\textwidth}
         \centering
         \includegraphics[width=\columnwidth]{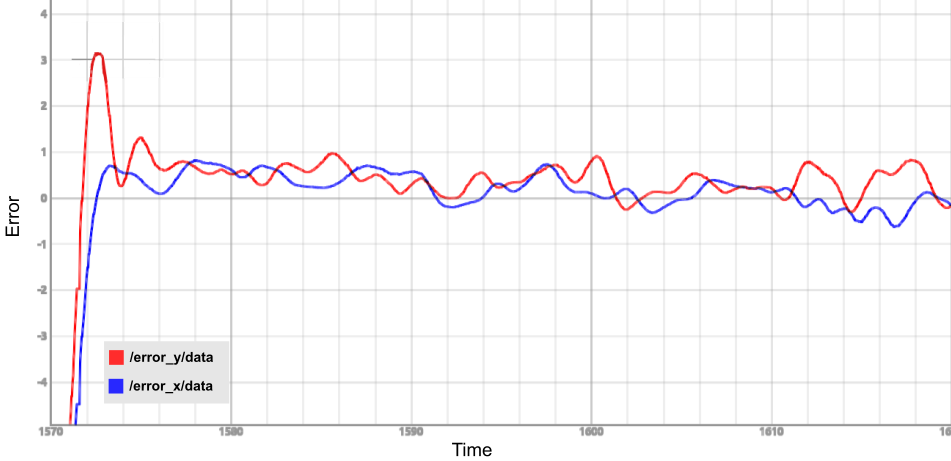}
         \caption{Some overshoot PID controller}
         \label{fig:less_overshoot}
     \end{subfigure}
     \hfill
     \begin{subfigure}{0.46\textwidth}
         \centering
         \includegraphics[width=\columnwidth]{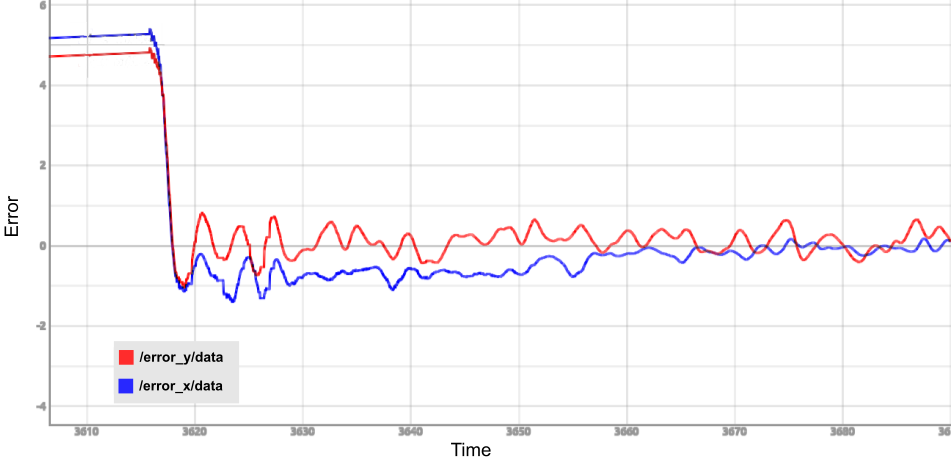}
         \caption{No overshoot PID controller}
         \label{fig:no_overshoot}
     \end{subfigure}     
     \hfill
        \caption{Plot of error in y (red) and x (blue) axes in WhyCon coordinate frame vs time in seconds for different types of PID controllers as per Ziegler-Nichols method.}
        \label{less_overshoot}
\end{figure}





\begin{table}[!b]
    \centering
\caption{Different types of controllers along with their parameters as per Ziegler-Nichols method}
\label{tab:zn_typesl}    
\begin{tabular}{|p{8em}|p{4em}|p{3.5em}|p{3em}|}\hline
		\textbf{Controller type} & \textbf{Kp} &\textbf{Ti} & \textbf{Td}\\\hline
		P  & 0.5 * Ku & $-$ & $-$\\\hline
			PI & 0.5 * Ku & Tu/1.25 & $-$\\\hline
PD & 0.8 * Ku & $-$ & Tu/8 \\\hline
Classic PID & 0.6 * Ku & Tu/2 & Tu/8\\\hline
	Pessen Integral rule & 0.7 * Ku & Tu/2.5 & 3Tu/20 \\\hline
Some overshoot & 0.33 * Ku & Tu/2 & Tu/3\\\hline
No overshoot & 0.2 * Ku & Tu/2 & Tu/3\\\hline

\end{tabular}

\end{table}

 We force the controller output to maximum and wait for the NAV to cross over the setpoint. On crossing the setpoint, the controller output is forced to a minimum. This is repeated four to five times to obtain approximately consistent oscillation about the setpoint. This process is repeated for pitch and then roll as well. We consider that the controller gain necessary to maintain this oscillation is $K_u$ and the time period between this oscillation is $T_u$. $K_u$ is computed as given in Equation \ref{eq:zn_ku},
\begin{equation}
    K_u=\frac{4\times{d}}{pi\times{a}}
    \label{eq:zn_ku}
\end{equation}
        where $d$ is the amplitude of PID output and $a$ is the amplitude of oscillation about the setpoint. The controller can be represented as given in Equation \ref{eq:zn_pid},



        \begin{equation}
        u(t) = K_p \bigg( e(t) + \frac {1}{T_i} \int_{0}^{t}e(t)dt + T_d \frac{de(t)}{dt}\bigg)
        \label{eq:zn_pid}
        \end{equation}
        where, $T_i$ is the integral time constant and $T_d$ is the derivative time constant. According to the Ziegler-Nichols method, Table \ref{tab:zn_typesl} establishes the relation between the above-mentioned parameters.

Depending upon the need, an appropriate controller type is selected and their corresponding equations can be used to compute the required parameters. From the values of $T_i$, $T_d$, and $K_p$ we deduce the values of $K_i$ and $K_d$ as per Equation \ref{eq:zn_ki}.
\begin{equation}
    K_i = \frac{K_p}{T_i},   K_d=K_p\times{T_d}
    \label{eq:zn_ki}
\end{equation} 
  

Once $K_p$, $K_i$, and $K_d$ are determined they can be used in the previously designed PID controller. Experimental results of different types of controllers tuned in the $x$ and $y$ axes are shown in Figure \ref{less_overshoot}.








\begin{figure}[!t] 
\centering
\includegraphics[width=0.8\linewidth]{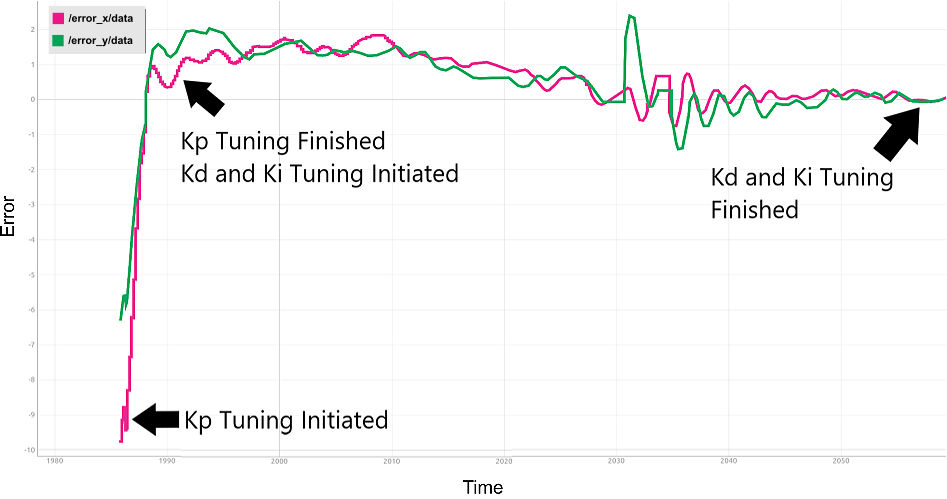}
\caption{Plot of error vs time in $x$ (pink) and $y$ (green) axes observed during auto-tuning of the controller as per Algorithm~\ref{iteration_algo}. The range of values entered by the user for $K_p$, $K_i$ \& $K_d$ of $\phi$, $\theta$ and $throttle$ are used as the starting point for determining the optimum values for these parameters.}.
\label{Iteration_Based_Autotuning_Graph}
\end{figure}

\subsubsection{Iterative Feedback Auto-tuning}
\label{Iterative Feedback Auto-tuning Section}

This method calculates optimum values of PID parameters of $\phi$, $\theta$ \& throttle for the controller by tuning during the flight. The user enters a range for the possible values of the parameters and the code continuously changes them based on Algorithm \ref{iteration_algo}. Figure \ref{Iteration_Based_Autotuning_Graph} shows the results obtained for tuning $x$ and $y$ axes based on Algorithm \ref{iteration_algo}.

\begin{algorithm}[ht]
\caption{Algorithm for Iterative Feedback Auto-tuning}
\label{iteration_algo}
\begin{algorithmic}[1]
\renewcommand{\algorithmicrequire}{\textbf{Input:}}
\renewcommand{\algorithmicensure}{\textbf{Output:}}
\REQUIRE Range of values for $K_p$, $K_i$ \& $K_d$ of $\phi$, $\theta$ \& $throttle$.
\ENSURE Tuned PID controller values.
\STATE Set $K_p$, $K_i$ \& $K_d$ of $\phi$ \& $\theta$ to minimum.
\STATE Set $K_p$ to maximum and $K_i$ \& $K_d$ of $throttle$ to minimum.
\WHILE {$err >= threshold$ in all axes}
    \STATE Increase $K_p$ for $\phi$ \& $\theta$ and decrease for $throttle$ by 0.1 units.
\ENDWHILE

\WHILE{$err >= threshold'$ in all axes}
    \IF {$(err_{max} - err_{min}) >= threshold$ in all axes}
        \IF {$overdamped$ in any axis}
            \STATE Decrease $K_d$ for the axis w.r.t. $f'(err)$.
        \ELSE
            \STATE Increase $K_d$ for the axis w.r.t. $f'(err)$.
        \ENDIF
    \ENDIF
    \IF {$err >= threshold$ in all axes}
        \STATE Increase $K_i$ by 0.5 units
        \IF {$overdamped$ in any axis}
            \STATE Decrease $K_d$ for the axis w.r.t. $f'(err)$.
        \ELSE
            \STATE Increase $K_d$ for the axis w.r.t. $f'(err)$.
        \ENDIF
    \ENDIF
\ENDWHILE

\end{algorithmic}
\end{algorithm}







\section{Applications on Proposed Setup}\label{application}


In this section, three applications are discussed where the proposed system works seamlessly. In subsection \ref{auto_landing}, a novel indoor autonomous landing of NAV on a mobile platform is implemented, in subsection \ref{path_plan}, indoor path planning of NAV while avoiding obstacles is implemented, and in subsection \ref{multi_drone}, control of multiple drones to create simple formations is implemented.

\subsection{Autonomous landing on moving platform}
\label{auto_landing}

Over the last few decades, numerous research groups have dedicated substantial efforts to investigating a variety of techniques for achieving autonomous drone landings. One approach, as outlined in \cite{autoLanding}, involves the application of color-based image processing algorithms. These algorithms empower drones to perform landings in diverse settings, encompassing both indoor and outdoor environments.

In contrast, other strategies, detailed in \cite{yang2013onboard} and \cite{olivares2015vision}, rely on onboard vision systems to identify specific symbols or markers, such as the 'H' enclosed in a circle, which serve as landing pads in indoor environments. Additionally, the fusion of optical flow sensors and marker detection techniques, as explored in \cite{lee2012autonomous} and \cite{li2011vision}, enhances the precision of landing, especially on moving objects.

Nonetheless, the majority of these methods depend on onboard vision-based detection of landing targets, which imposes limitations on the camera's field of view (FoV). The fixed FoV, as presented in the proposed methodology, offers an effective solution for autonomous landing on moving objects. It's worth noting that this technique is most applicable within controlled environments. However, it still holds value as a testing and evaluation platform for algorithms designed to facilitate drone landings on moving targets. In the subsequent section, we will delve into the setup and algorithm for executing autonomous landings on moving targets, as per the proposed methodology.

\begin{figure}[!t] 
\centering
\includegraphics[width=0.7\linewidth]{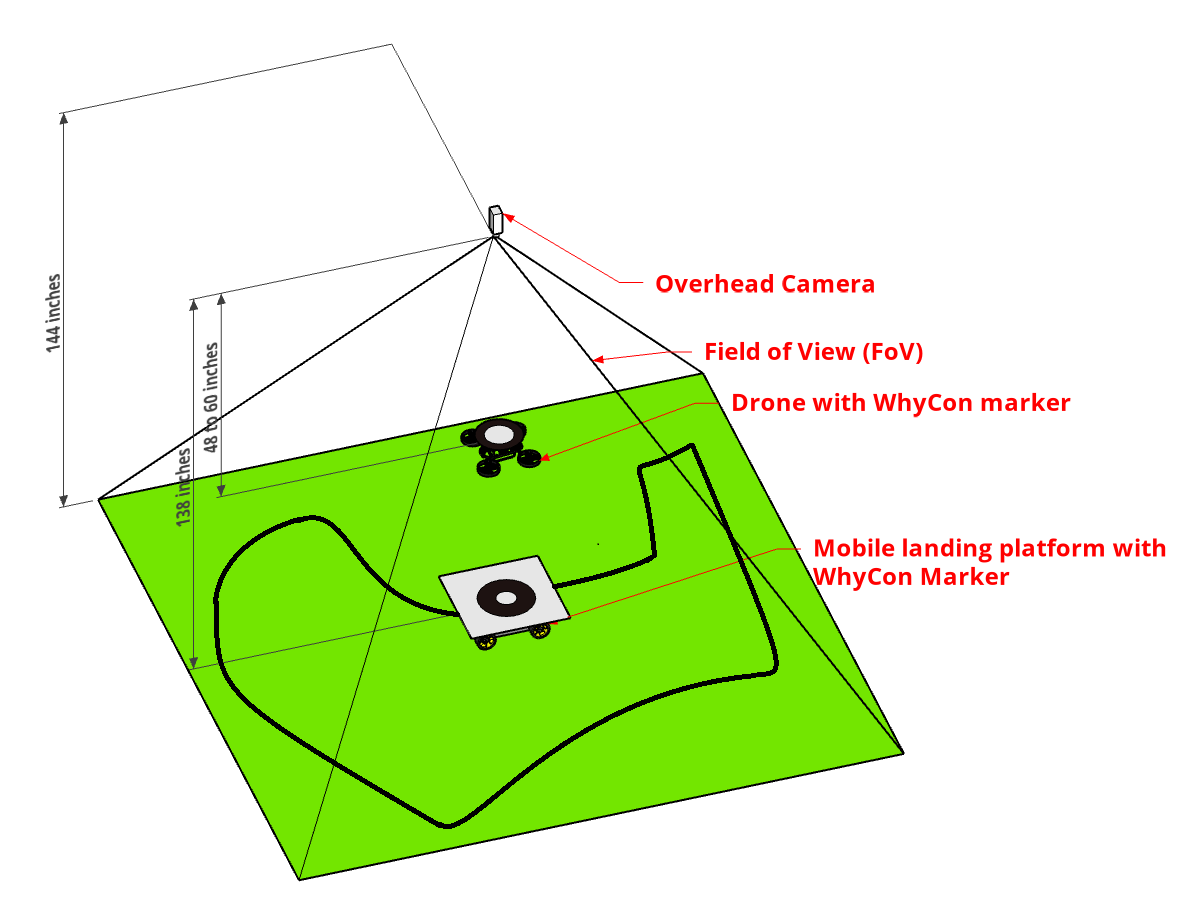}
\caption{Experimental setup for autonomous landing on a moving platform.}
\label{cd_setup}
\end{figure}

\subsubsection{Setup}

The entire setup is depicted in Figure~\ref{cd_setup}. An overhead camera with a fixed Field of View (FoV) is positioned as indicated in the figure. WhyCon markers are affixed to both the NAV and the mobile target. In real-time, the overhead camera employs a modified version of the WhyCon tracking algorithm (for details of the modification, see section \ref{auto_lan_ctrl_algo}) to track these markers. While the mobile target can navigate any path within the controlled environment, its motion is confined to two-dimensional space. As explained in section \ref{Internal and External Control}, the NAV relies on the overhead camera's perspective and off-board processing, as its onboard sensors are unable to self-localize. However, the overhead camera effectively tracks the mobile target and issues landing commands to the NAV to ensure a successful landing on the mobile target.

The mobile target in this setup consists of a NEX Robotics differential drive robot~\cite{Nex}, an embedded platform equipped with an Atmel AVR ATmega16 microcontroller. To facilitate tracking within the camera's Field of View (FoV), a WhyCon marker is mounted on the robot. To constrain the robot's movement within the controlled environment, a flexible sheet with a printed black line is used. This allows the mobile target to move freely in any direction while using the black line as a reference. The localization system, described in Section \ref{enviornment_setup}, is employed in this study to determine the $3D$ pose and heading angle of the NAV.

\begin{figure}[!t] 
\centering
\includegraphics[width=0.9\linewidth]{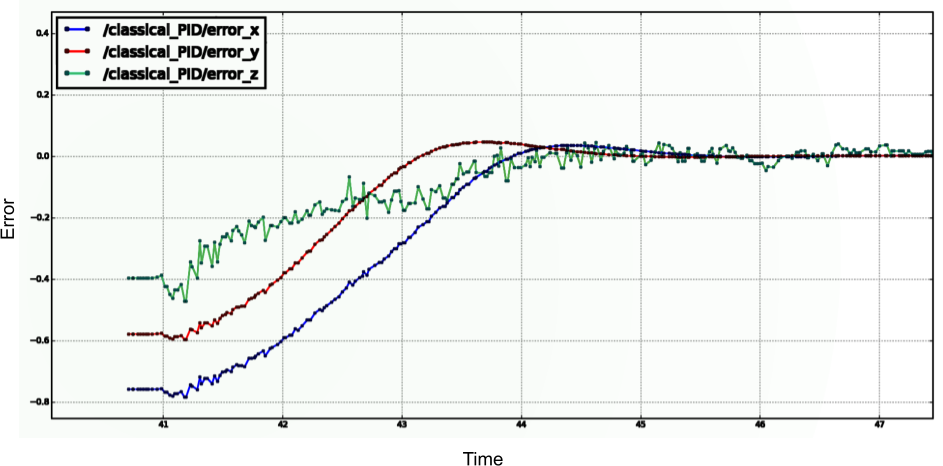}.
\caption{Plot of error in x (blue), y (red), and z (green) axes vs. time for the PID algorithm used in the simulation. The graph depicts that all three errors converge close to $0.0$ which is well within our desired accuracy.}
\label{CD_PID graph}
\end{figure}

\subsubsection{Controller and Algorithm}\label{auto_lan_ctrl_algo}
To achieve an autonomous landing of the NAV on a moving target, it is crucial for the NAV to be aware of the 2D location of the moving target. In our case, this information is provided by the WhyCon marker mounted on the mobile landing platform, which serves as the target's location (referred to as the set point in this paper). However, due to the random path followed by the target, the changing set point introduces a nonlinear error during the error calculation interval known as the loop time. This nonlinear error, which is directly influenced by the speed of the platform, is determined at each loop time. The varying speed affects the response of the PID; hence, for simplicity within this experiment, the mobile platform was kept at a constant speed.

The proposed algorithm requires tracking two targets at a time. It can be quickly done in the existing WhyCon package. However, this particular tracking algorithm should switch the target tracking objects in FoV at run time for efficient tracking. We modified the existing WhyCon library to handle the run-time target change. According to the modification, the algorithm iterated for the maximum number (`n') of the target defined during initialization. If the desired number of markers are not detected in the frame, it issues a service and starts tracking again with an n-1 target in FoV.


Hence for our application, the predefined number of markers is 2. When the drone hovers/flies over the moving platform, the WhyCon marker detection algorithm in the library fails to detect both markers, which causes the detection algorithm to stop working momentarily. This instability for the detection algorithm affects the PID control algorithm. A marker coordinate feedback was implemented in the control algorithm to solve this detection algorithm problem since in our application, the $z$ coordinate of the drone is always less than that of the mobile platform as illustrated in Figure \ref{cd_setup}. Every time the detection algorithm stops returning the coordinates of the markers, our service restarts the tracking and changes the predefined value from 2 to 1.


To test the algorithms and environment, we used the Gazebo simulator and the modified Ardrone 2.0 model from tum\_simulator. The model featured the WhyCon marker affixed to it. The PID response during the simulation was recorded and the graph is displayed in Figure \ref{CD_PID graph}.

\subsection{Path planning and autonomous traversal}
\label{path_plan}
Several authors have sought to develop a system for path planning and autonomous traversal of indoor drones. Researchers in \cite{indoorracing} have proposed a deep-learning and Line-Of-Sight (LOS) guidance algorithm for autonomous indoor navigation of drones. Others \cite{scalprecmultiuav} have deployed receiver-side time-difference of arrival (TDOA) based ultra-wideband (UWB) indoor localization for drones. The authors of \cite{lowcostquad, sani2017automatic} have used a camera and Inertial Measurement Unit (IMU) to navigate the drone. The Swarm of Micro-Flying Robots (SFLY) project \cite{visionmicro} uses monocular simultaneous localization and mapping (SLAM) fused with inertial measurements to navigate the drone in a GPS-denied environment. The work described in \cite{viconparrot} concludes that the position estimation of the Vicon camera is better than the onboard sensors' estimation. In search of a low-cost localization system, our application involves the autonomous traversal of a NAV through hoops using WhyCon.

\begin{figure}[!t] 
\centering
\includegraphics[width=\linewidth]{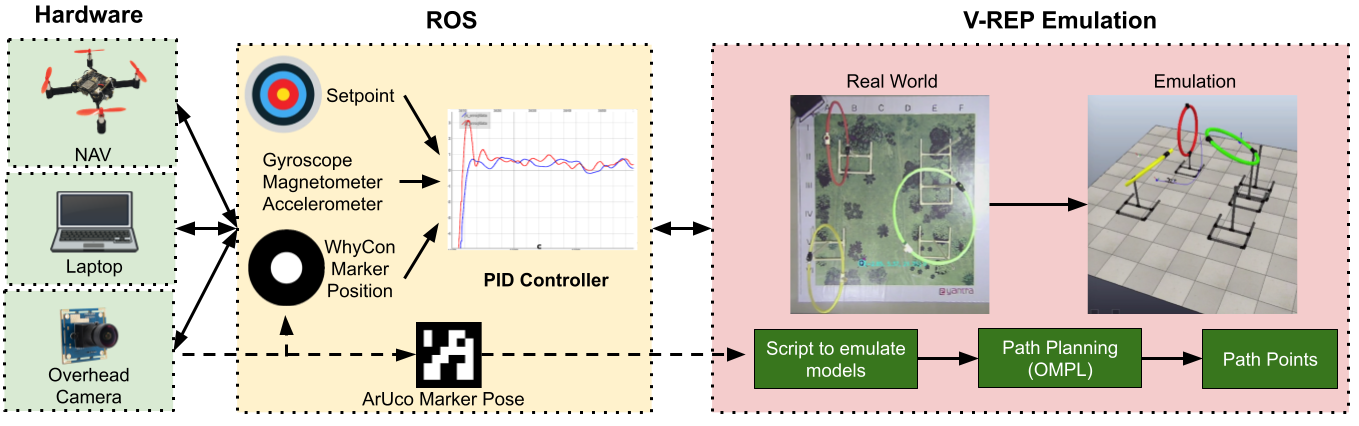}
\caption{System architecture for path planning and autonomous traversal.}
\label{System_architecture}
\end{figure}

Using three-dimensional path planning, an optimal and collision-free path is found considering various kinematic constraints. Three-dimensional path planning algorithms include Probabilistic Road Maps, bio-inspired planning algorithms, optimal search algorithms like A* \& Dijkstra's algorithm, and random-exploring algorithms like RRT and RRT*. Sampling-based algorithms like RRT offer high time efficiency and are more suitable for real-time implementation \cite{yang2016survey}. In our work, we use RRT* path planning algorithm because of the higher probability of converging to an optimal solution compared to RRT when obstacles are present \cite{karaman2010incremental}. RRT* simply builds a search tree of reachable states by attempting to apply random actions at known-reachable states. The action is considered successful if it does not cause any contact with any obstacle and the resulting state is added to the tree of reachable states \cite{lavalle2001randomized}. Implementation of these sampling algorithms is available in Open Motion Planning Library (OMPL). All these planners operate on very abstractly defined state spaces. OMPL plugin is available in many simulation software. In our work, we have used the V-REP simulator (now CoppeliaSim). V-REP has multiple physics engines, mesh manipulation and most importantly it provides the user the ability to interact with the world during the simulation run. Compared to Gazebo, V-REP is more intuitive, user-friendly, and consumes less CPU power \cite{nogueira2014comparative}. Hence in our work, we are using V-REP as the simulation environment to test the method and later validate the method with a real NAV.


\begin{figure}[!t] 
\centering
\includegraphics[width=0.8\columnwidth]{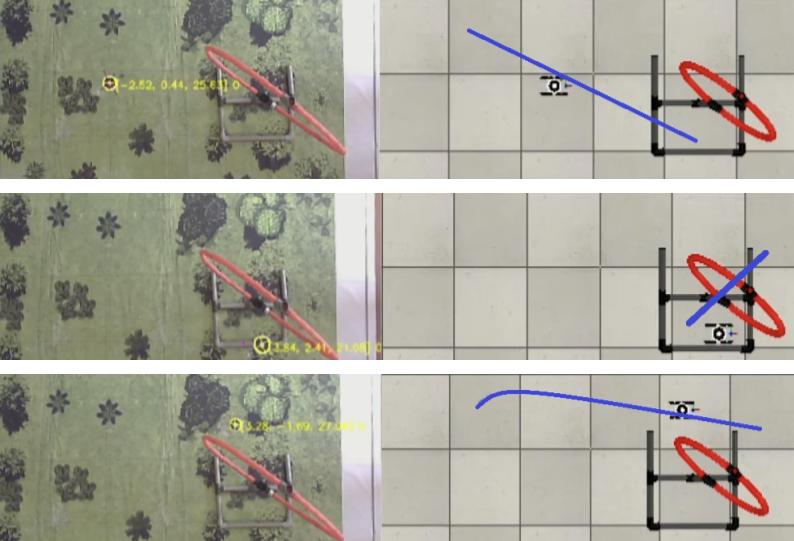}
        \caption{Set of three paths followed by NAV for traversing through a hoop. WhyCon coordinates and path to be traversed are shown in yellow and blue color respectively. \underline{Top to bottom}: NAV moving towards the entry location of hoop, NAV passing through the hoop and NAV moving away from the hoop.}
        \label{fig:drone_traversal_through_hoop}
\end{figure}

\subsubsection{Implementation of the project in V-REP}
Testing the algorithm in V-REP involves the modeling of NAV and hoops in V-REP similar to the ones in the real world, making a V-REP scene for the experimental setup and the implementation of the algorithm. Calculation of path is done in V-REP using OMPL library. Path points are being published as the required waypoints which are subscribed by the PID controller node in ROS to direct the NAV as shown in Figure \ref{System_architecture}. 


\subsubsection{Modeling of NAV in V-REP}

A model is designed in V-REP which subscribes to the same topic of the same message type as the real NAV. The model is controlled by ROS commands. The model publishes its orientation with respect to the V-REP world. It subscribes to the rostopic \textit{drone\_command}. For eg: To disarm the drone following command is to be published:\\ \\ \textbf{rostopic pub /drone\_command pluto drone/PlutoMsg "\{rcRoll: 1500,
rcPitch: 1500, rcYaw: 1500, rcThrottle: 1000, rcAUX1: 0, rcAUX2:
0, rcAUX3: 0, rcAUX4: 1200\}"} 



\subsubsection{Setting up an experimental scene in V-REP}
A scene is setup in V-REP comprising of overhead vision sensor and hoops as shown in the Emulation block of Figure \ref{System_architecture}. On running the simulation, the vision sensor publishes images of a resolution of 640 x 480 at 30 frames per second. Localization of NAV is done using WhyCon markers. WhyCon node subscribes to the image and publishes relevant output topics.



\begin{figure}[!t] 
\centering
\includegraphics[width=0.5\linewidth]{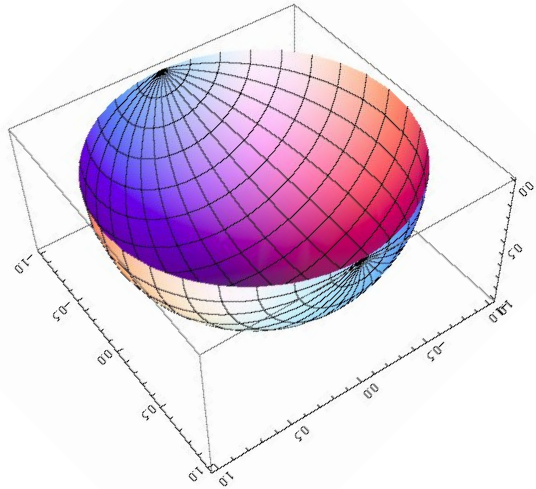}
\caption{Representation of z-coordinate readings of WhyCon marker along the $x$ and $y$ axis.}
\label{z_coordinate}
\end{figure}

\subsubsection{Computation of path using OMPL}

The Open Motion Planning Library, OMPL consists of many sampling-based motion planning algorithms such as PRM, KPIECE, EST, SBL, RRT and operates on defined state spaces. Path and motion planning can be done in V-REP using a plugin wrapping the OMPL library. Path planning involves creating a path planning task, creating the required state space, specifying collision pairs, setting start, and goal states, and selection of path planning algorithm followed by computing the required path. RRT* is the path planning algorithm used here. Multiple paths are computed between the start \& goal state and the shortest path is selected for the NAV to navigate. A minimum of fifty path points are made between the NAV \& goal state and these path points are being published to the NAV as new waypoints to follow. As the nano drone's flight time is of 7 minutes, a maximum time of 20 seconds has been set for each path-searching procedure. Traversal through one hoop involves finding three paths as shown in Figure~\ref{fig:drone_traversal_through_hoop}. In the worst case where each path takes 20 seconds to plan and there are three such hoops, it will take three minutes for planning all the paths for all three hoops. The remaining flight time is enough for the drone to maneuver the path.

\subsubsection{Emulation of real-world in V-REP}
Emulation involves the process of replicating the real world into the V-REP environment so that the computed path points can be directly given as waypoints for NAV to navigate. Input from the WhyCon marker is used to emulate the NAV model. Input from the ArUco markers \cite{arucoopencv} is used to emulate the pose of the hoops. An ArUco marker is a square marker composed of a wide black border and an inner binary matrix that determines its identifier. ArUco marker gives the pose of the marker with respect to the camera. After emulating the hoops and obstacles ArUco markers are removed as the environment is static.


\subsubsection{Mapping between WhyCon frame and V-REP frame}
\label{map_whycon_vrep_frame}
For a given WhyCon marker with an outer diameter of 0.055m and inner diameter of 0.02m, one unit of x and y coordinate approximately equals 10 cm in real or V-REP world. However, even if we keep the marker at the same height, the value of WhyCon's z-coordinate is not constant when we move the marker along the $x$ or $y$ axis. We observed the trend in the change of the z-coordinate similar to a semi-ellipsoid, as shown in Figure \ref{z_coordinate}. To minimize this error we used the equation of an ellipsoid to fit the z-coordinate. WhyCon readings are hence converted to the real world or V-REP world using corresponding scaling factors.

\subsubsection{Traversal of NAV through hoops}
After converting path points in the V-REP frame to the WhyCon frame of reference, each point is given as a waypoint for the NAV to navigate. The emulation block in Figure \ref{System_architecture} shows the traversal of NAV through hoops.


\begin{table}[t!]
\centering
\caption{Comparison of the existing systems to our system}
\label{tab:my-table}
\begin{tabular}{|l|c|c|c|c|}
\hline
\textbf{Research} &
  \textbf{\begin{tabular}[c]{@{}c@{}}Technologies \\ Used\end{tabular}} &
  \textbf{\begin{tabular}[c]{@{}c@{}}Localization \\ Error (cm)\end{tabular}} &
  \textbf{\begin{tabular}[c]{@{}c@{}}Positioning \\ Rate (Hz)\end{tabular}} &
  \textbf{\begin{tabular}[c]{@{}c@{}}Cost(excl.\\ 
 drone)\end{tabular}} \\ \hline
\begin{tabular}[c]{@{}l@{}}Paredes, J.A. \\ et al.\end{tabular} &
  \begin{tabular}[c]{@{}l@{}}Ultrasonic and \\ TOF cameras\end{tabular} &
  4 – 8 &
  2 &
  \textless 100 USD \\ \hline
\begin{tabular}[c]{@{}l@{}}Khalaf-Allah, \\ M. et al.\end{tabular} &
  \begin{tabular}[c]{@{}l@{}}Ultra wide band \\ Tags and Anchors\end{tabular} &
  26 &
  50-70 &
  \textless 300 USD \\ \hline
\begin{tabular}[c]{@{}l@{}}Nenchoo, B. \\ et al.\end{tabular} &
  \begin{tabular}[c]{@{}l@{}}Depth Camera \\ and IMU\end{tabular} &
  10 &
  10 &
  \textless 150 USD \\ \hline
\begin{tabular}[c]{@{}l@{}}Tiemann, J \\ et al.\end{tabular} &
  \begin{tabular}[c]{@{}l@{}}Ultra wide band \\ and vSLAM\end{tabular} &
  14 &
  32 &
  \textless 400 USD \\ \hline
\begin{tabular}[c]{@{}l@{}}Tiemann, J \\ et al.\end{tabular} &
  \begin{tabular}[c]{@{}l@{}}Ultra wide band, \\ Camera, IMU\end{tabular} &
  26 &
  40 &
  \textless 300 USD \\ \hline
\textbf{Our System} &
  \textbf{RGB Camera} &
  \textbf{3 - 4} &
  \textbf{4} &
  \textbf{\textless 50 USD} \\ \hline
\end{tabular}
\end{table}

\subsection{Multi-drone control}
\label{multi_drone}
Swarm drones/robots have various applications, both indoor and outdoor. This application focuses on multiple drone control in the indoor environment. When indoor multi-drone localization is involved, sufficient research has been done. Vanhie-Van, G. et al. \cite{vanhie2021indoor} have summarized the sensor and sensor fusion technique used in multi-drone localization. The study mainly covered three classes of drones: large, medium, and nano. When the authors analyzed the cost, the sensor cost alone came to ~400 USD; in our case, the cost of drones and sensors comes below 100 USD. As seen in Table  \ref{tab:my-table}, there are a few trade-offs, but our system works better for similar applications and use cases.

In our application, drones perform coordinated motion where each drone follows a specific path relative to other drones within a network. Communication delay has also been calculated and discussed for multiple coordinated tasks in Section \ref{multi_drone_latency}. Each drone in the network acts as a client. We used ROS to implement a swarm of drones. In ROS, multiple nodes are created, which are required for communication, localization, and control of formations. Each drone is assigned a unique topic within the network. Drone commands are published on this topic while each drone subscribes to its unique topic and sets its parameters accordingly. The techniques discussed in the previous implementation \ref{auto_landing} were used for localization. 

\begin{figure}[!t] 
     \centering
     \begin{subfigure}{0.6\textwidth}
         \centering
         \includegraphics[width=\columnwidth]{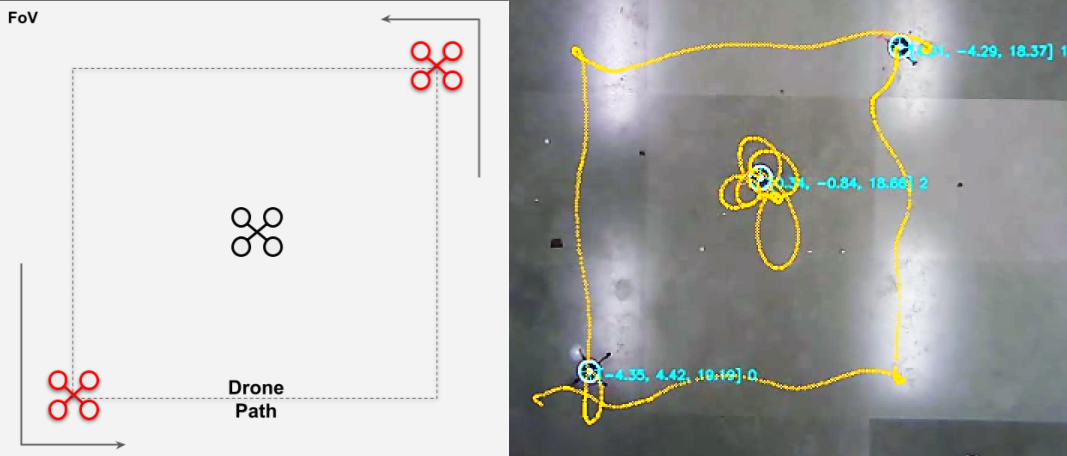}
         \caption{Square traversal by three NAVs}
         \label{fig:Square traversal by two NAVs}
     \end{subfigure}
     \hfill
     \begin{subfigure}{0.6\textwidth}
         \centering
         \includegraphics[width=\columnwidth]{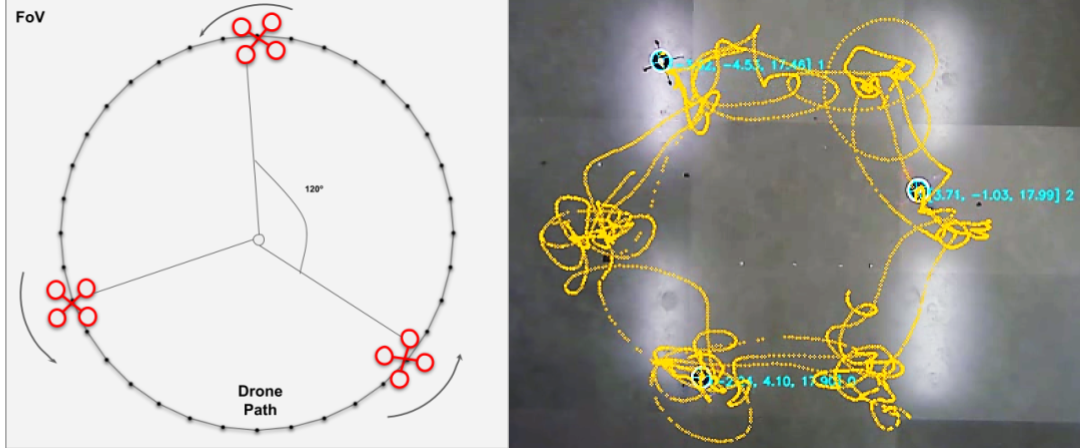}
         \caption{Circular traversal by three NAVs}
         \label{fig:Circular traversal by three NAVs}
     \end{subfigure}
     \vfill
        \caption{Traversal of different shapes by NAVs}
        \label{fig:three graphs}
\end{figure}

\subsubsection{Communication and Feedback Control}

Each drone is connected to an access point. Once connected, the IP address of each drone is identified and marked. A laptop running ROS is connected to the same access point. Individual threads are implemented in C++ to communicate with each drone separately with their unique socket. Using these individual sockets, operations like read and write with each drone were achieved. Sensor data of each drone is published on individual topics through the implemented sockets. The PID control system is implemented on individual threads for each drone. Each thread reads the position of a particular drone and calculates $\phi$, $\theta$, $\psi$, and throttle for that drone for a fixed target position. At start-up, each thread is initialized for its associated drone with its PID values and drone number as an argument. The drone number is associated with the IP address of the drone.

\subsubsection{Distinguishing between multiple markers}
Each drone has an identical WhyCon marker for localization with an id assigned to it by the WhyCon package. Since all drones had relatively close $z$ values, they could not be used as a distinguishing factor between the markers. To overcome this and to avoid a mix-up of IDs, each drone is assigned an approximate fixed position at start-up according to the IP address of the drone. Once the script is started, the previous position of each drone is used to track the drone individually, irrespective of the position of the drone. Marker values are compared with the previous marker position, and correct values are assigned to the global list depending on the error between previous and current positions. This method, however, creates latency and is discussed in subsection \ref{multi_drone_latency}.

\begin{table}[!t]
\centering
\caption{Overall latency in multi-drone control system}
\begin{center}
\begin{tabular}{|c|c|}
\hline
    Case & latency in ms  \\
\hline
    Camera frame capture     & 8.33 \\
    Marker detection  &  12 \\
    Drone Identification  &  27 \\
    PID Loop  &  100 \\
    MSP packet  &  190 \\
    Communication  &  4 \\
    \hline
    Total & 341.33 \\
     
\hline
\end{tabular}
\label{tab1:latency}
\end{center}
\end{table}
 
\subsubsection{Formation and traversal of drone}
To achieve the desired formations, this paper describes the methodology used to update the target coordinates of each drone. This involves taking feedback from the drones' positions within FoV. By continuously monitoring and analyzing the relative positions of the drones, adjustments to the target coordinates are made to ensure that the desired formation is maintained. In order to evaluate our multi-drone control, circular \& square traversal and synchronized circular three-drone formation was implemented using three drones as shown in Figure \ref{fig:three graphs}

\textbf{Circular traversal:} The set point of the drone is changed continuously by shifting every 10$^\circ$ to create an arc. The drone completes a circular trajectory in 36 set point coordinates. The arc coordinates are at 10$^\circ$, and the difference is calculated as shown in Equation \ref{eq: circle traversal}, where $a$ ranges from 0 to 36 and $r$ are the circle radius. $dist\_apart$ and $wc\_id$ are formation control variables set to zero for circular traversal. 

\begin{equation}
\begin{split}
    x = r\times sin((a\times10) + (dist\_apart\times wc\_id)) \\
    y = r\times cos((a\times10) + (dist\_apart\times wc\_id))
\end{split}
\label{eq: circle traversal}
\end{equation}
   
\textbf{Square traversal:} This traversal is implemented by calculating the vertices of a square of a predefined size. Two drones opposite each other traverse the vertices to form a square. As shown in Figure \ref{fig:Square traversal by two NAVs}, one drone holds its position at the center.

\textbf{Rotation formation}: Rotation by three drones is achieved in the same way as circular traversal. However, a 120$^\circ$ arc is created for each drone as shown in Figure \ref{fig:Circular traversal by three NAVs}. The new coordinates are calculated as shown in Equation \ref{eq: circle traversal} by setting $dist\_apart$ to 120$^\circ$ and $wc\_id$ (Whycon ID) as 0,1,2 for our rotation formation.

\subsubsection{Latency}
\label{multi_drone_latency}
The different factors which cause latency in the multi-drone control system are calculated and presented in Table~\ref{tab1:latency}. The brief description of calculated latency is as follows:

\noindent \textbf{\textit{Marker detection latency}:}
The image is processed using the WhyCon package to detect the markers in the frame. The package uses the OpenCV library to detect the marker using the flood fill algorithm. It takes an average of 12 milliseconds to detect the marker and determine the relative coordinates of the marker within the frame.\\
\textbf{\textit{Latency in drone identification}:}
Every drone has the same pattern of markers; hence it is challenging to keep track of individual drones when they hover around within the FoV. The initial position of the drone is fixed in the frame. The previous position of the drone is saved in every iteration to calculate the nearest previous position to determine the current positions of the drones. It takes 27 milliseconds to calculate the positions of each drone.\\
\textbf{\textit{Feedback control loop latency}:
}Parallel PID is used to control the drone's $\phi$, $\theta$, $\psi$, and throttle. The feedback control loop for each drone takes 100 milliseconds\\
\textbf{\textit{Command packet latency}:}
After calculating $\phi$, $\theta$, $\psi$, and throttle by a feedback control loop, this data is converted to an MSP packet for sending through the network. The firmware on the drone receives the MSP packet and decodes it according to the MSP protocol. The conversion from raw data to MSP packet takes 190 milliseconds.\\
\textbf{\textit{Latency in communication}:
}The communication medium is wireless, and the topology used is star network, i.e., MSP packets first go to the router and then routed to the drone. Hence the communication latency is 4 milliseconds.\\
\textbf{\textit{Thread scheduling latency}:}
Every thread has some clock assigned by the scheduling algorithm. The frequency of the time slot assigned to a thread depends on the number of threads running and each thread's requirement and priority. Hence, the latency due to each process can vary due to thread scheduling. Variance in latency in the feedback control thread is directly proportional to scheduling.

\subsubsection{Challenges faced}
Differentiating between the drone ID was overcome by comparing the markers' current position with the previous position of the marker. Individual threads were implemented for the PID control of each drone, along with tuned PID constants of the drone within that thread. The most daunting challenge faced was the scheduling of threads. An operating system assigns time slots to different threads according to available resources. This creates an issue as control threads calculation must be done in real-time. This problem can be solved by using a real-time operating system or by decreasing the number of threads such that the threads get time slots frequently.

\section{Conclusions}\label{conclude}

This paper presents a localization approach to control the NAV in an indoor environment using a monocular camera. This approach allows controlling the NAV in four degrees of freedom with a maximum latency of approximately 0.341 seconds and an average localization error of 3.1 cm. The package developed for communicating between NAVs and ROS systems ensures minimum latency for control commands. Multiple parallel proportional-integral-derivative controllers helped to stabilize the drone in the custom-controlled environment. The applications developed on the proposed techniques show the various possibilities suitable for a low-cost academic environment. Autonomous path planning indicates an affordable and accurate means to traverse the drone while avoiding obstacles. Landing of NAV on a moving platform and multi-drone control demonstrates the scope to use numerous components in our approach. We can even establish this localization system in an indoor environment like a warehouse to automate various robotic systems like drones and differential drive robots. In the future, we plan to scale the system with multiple monocular cameras to expand the controlled environment.

\section*{Acknowledgement(s)}

We would like to thank Ministry of Education (MoE), Govt. of India and Prasanna Shevare, CTO of Drona Aviation, \& Krishna Zore, Senior Software Engineer of Drona Aviation, for their support with the NAV.

\section*{Disclosure statement}

On behalf of all authors, the corresponding author states that there is no conflict of interest.

\section*{Funding}

This work was funded by the Ministry of Education (MoE), Govt. of India, in the project e-Yantra (RD/0121-MHRD000-002).

\section*{Data availability statement}
The data that support the findings of this study are available from the corresponding author upon reasonable request. The ROS package developed to maneuver the NAV in the controlled environment is openly available at \\ 
\url{https://github.com/simmubhangu/eyantra_drone}

\section*{Notes on contributor(s)}

\textbf{Simranjeet Singh} received a B.Tech degree in electronics and communication from Kurukshetra University, Kurukshetra in 2015. He is working towards an M.Tech.+Ph.D. in the Department of Electrical Engineering at IIT Bombay, India. His main research area includes hardware security, embedded systems, and neuromorphic computing.\\

\noindent \textbf{Amit Kumar} received a B.E degree in Electronics from the University of Mumbai, India in 2020. He is currently working as a project staff in the ERTS/e-Yantra Lab of the Computer Science and Engineering department at IIT Bombay, India. His main research areas include Embedded Systems, and Field, \& Aerial Robotics. \\

\noindent \textbf{Fayyaz Pocker Chemban} received a B.Tech degree in Mechanical Engineering from TKM College of Engineering, Kerala in 2015. He is currently working as Senior Robotics Engineer at Turf Tank, Denmark leading the development of navigation stack on an automatic line marking robot.\\

\noindent \textbf{Vikrant Fernandes} is a Co-Founder at Physique Health Studio where is he currently building an AI powered fitness app called FitYoga. Vikrant is a Product Developer at heart and enjoys building experiences for end users.\\

\noindent \textbf{Lohit Penubaku} received a MSEE and MSES from Louisiana State University, USA. He is currently working as a project staff in the ERTS/e-Yantra Lab and a Research Scholar in Department of Electrical Engineering, IIT Bombay, India. His main research area includes Embedded Systems, Precision Agriculture, Edge ML and Remote Sensing.\\

\noindent \textbf{Kavi Arya, D.Phil.(Oxon.)} is Professor of Computer Science \& Engineering at the Indian Institute of Technology (Bombay). He is the Principal Investigator of the e-Yantra Project \url{https://www.e-yantra.org}, funded by the Ministry of Education (Govt. of India) and hosted at IIT Bombay.\\

\bibliographystyle{tfnlm}
\bibliography{interactnlmsample}
\end{document}